\documentclass[11pt]{article}

\usepackage{acl}
\pdfoutput=1

\usepackage{times}
\usepackage{latexsym}

\usepackage[T1]{fontenc}

\usepackage{placeins}  
\usepackage[utf8]{inputenc}

\usepackage{microtype}

\usepackage{inconsolata}

\usepackage{graphicx}

\usepackage{CJKutf8}

\usepackage{booktabs} 
\usepackage{multirow}
\usepackage{adjustbox}
\usepackage[table]{xcolor}
\usepackage{enumitem}
\usepackage{wrapfig}
\usepackage{amsmath}  
\usepackage{amsfonts} 
\usepackage[most]{tcolorbox}
\usepackage[utf8]{inputenc}
\usepackage[T1]{fontenc}    
\usepackage{xcolor}   
\usepackage{tikz}
\usetikzlibrary{tikzmark} 
\usepackage{fontawesome} 

\makeatletter
\newcommand*\myfontsize{%
  \@setfontsize\myfontsize{7}{8}%
}
\makeatother
\newcommand{\mytextbox}[2]{%
  \tikzmarknode[draw=#1,thick,inner sep=2pt]{mybox}{\myfontsize #2}%
}
\definecolor{mypurple}{RGB}{217, 208, 227}
\definecolor{myblue}{RGB}{0, 176, 241}
\definecolor{mygreen}{RGB}{17, 162, 62}

\newcommand{\blue}[1]{\mytextbox{myblue}{\textbf{\textcolor{myblue}{#1}}}}

\definecolor{gaincolor}{RGB}{0, 120, 60}  
\definecolor{losscolor}{RGB}{180, 0, 0}  

\newcommand{\res}[2]{%
  \ifdim #2 pt = 0.0 pt%
    #1%
  \else%
    \ifdim #2 pt > 0 pt%
       #1\rlap{\,\scriptsize$_{\textcolor{gaincolor}{+#2}}$}%
    \else%
       #1\rlap{\,\scriptsize$_{\textcolor{losscolor}{#2}}$}%
    \fi%
  \fi%
}   

\definecolor{myblue2}{RGB}{108, 140, 190} 
\definecolor{mybg}{RGB}{245, 248, 250}

\tcbset{
    promptstyle/.style={
        enhanced,
        colback=white,     
        colframe=myblue2,   
        coltitle=white,    
        fonttitle=\bfseries\large,
        boxrule=0.5mm,
        arc=2mm,
        left=6pt, right=6pt, top=6pt, bottom=6pt,
        drop shadow southeast,
        title={\faInfoCircle\hspace{0.5em}#1}
    }
}

\title{Analyzing Reasoning Consistency in Large Multimodal Models under Cross-Modal Conflicts}

\author{
    Zhihao Zhu$^{1}$\footnotemark[1], 
    Jiafeng Liang$^{1}$\thanks{$\,$ Equal Contribution.},
    Shixin Jiang$^{1}$,
    Jinlan Fu$^{3}$\footnotemark[2], 
    Ming Liu$^{1,2}$\thanks{$\,$ Corresponding Author.},
    \\
    \textbf{
    Guanglu Sun$^{1}$, See-Kiong Ng$^{3}$, Bing Qin$^{1,2}$
    } \\
    $^{1}$Harbin Institute of Technology, Harbin, China \\
    $^{2}$Peng Cheng Laboratory, Shenzhen, China \\
    $^{3}$National University of Singapore, Singapore \\
    \texttt{\{zhzhu,jfliang,sxjiang,mliu\}@ir.hit.edu.cn}
}

\begin{document}
\maketitle
\begin{abstract}
Large Multimodal Models (LMMs) have demonstrated impressive capabilities in video reasoning via Chain-of-Thought (CoT). However, the robustness of their reasoning chains remains questionable. In this paper, we identify a critical failure mode termed textual inertia, where once a textual hallucination occurs in the thinking process, models tend to blindly adhere to the erroneous text while neglecting conflicting visual evidence. To systematically investigate this, we propose the \textbf{LogicGraph Perturbation Protocol} that structurally injects perturbations into the reasoning chains of diverse LMMs spanning both native reasoning architectures and prompt-driven paradigms to evaluate their self-reflection capabilities.
The results reveal that models successfully self-correct in less than 10\% of cases and predominantly succumb to blind textual error propagation.
To mitigate this, we introduce \textbf{Active Visual-Context Refinement}, a training-free inference paradigm which orchestrates an active visual re-grounding mechanism to enforce fine-grained verification coupled with an adaptive context refinement strategy to summarize and denoise the reasoning history.
Experiments demonstrate that our approach significantly stifles hallucination propagation and enhances reasoning robustness.

\end{abstract}
\section{Introduction}
\begin{figure}[t]
    \centering
    \includegraphics[width=1\linewidth]{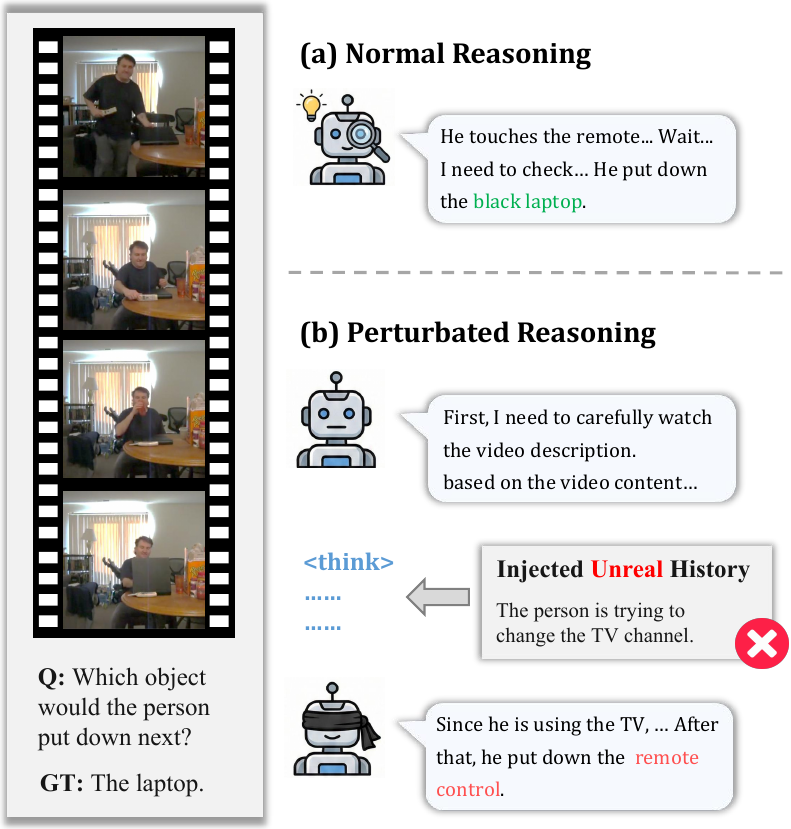}
  \caption{Illustration of visual blindness induced by erroneous textual context. While normal reasoning (a) grounds answers in visual evidence, perturbated reasoning (b) demonstrates that injecting a factual error causes the model to ignore conflicting visual signals. The model prioritizes consistency with the false history over visual reality, leading to incorrect justifications.}
    \label{figure:introduction_case}
\end{figure}

Large Multimodal Models (LMMs) ~\cite{Qwen2.5-VL,Keye-Pre,InternVL3} have demonstrated impressive capabilities in general video comprehension, evolving from simple perception~\cite{anqa} to complex reasoning tasks~\cite{STAR,videomme}. 
Unlike static image analysis, video reasoning requires models to maintain logical consistency across temporal sequences and comprehensively process spatiotemporal information correlations among multiple frames. 
Therefore, the ability to automatically reflect, verify and correct errors during the reasoning process has become an important research hotspot~\cite{videor1,open-r1-video}.

In the text-only domain, recent works like SCoRe~\cite{SCoRE} have successfully trained Large Language Models (LLMs) to self-correct via reinforcement learning, demonstrating that models can refine their outputs using self-generated data. Extending this to the multimodal sphere, Subsequent studies~\cite{SImRe,Volcano} further validate that LMMs also possess such reflective capabilities, enabling them to self-improve reasoning by explicitly reflecting on their own rationales.

However, a crucial question about reflective sources has been largely overlooked: \textit{When correcting reasoning steps, it remains unclear whether LMMs actively re-examine visual content or simply rely on history textual context.}

Motivated by this, we construct a preliminary analysis where we inject subtle factual errors into the early steps of a reasoning chain.  
We find that once a textual error is generated or injected, the model exhibits an overwhelming tendency to trust its own erroneous history over the conflicting visual evidence (shown in \autoref{figure:introduction_case}). 
Specifically, instead of looking back at the video to verify the facts, the model creates a justification based on the previous text, leading to a cascade of errors. 
This suggests that in current LMMs, the strong probability distribution of the language decoder often overrides the visual signal, rendering the model effectively blind during the reflection process.

To rigorously quantify this phenomenon, we introduce the LogicGraph Perturbation Protocol. Instead of treating reasoning as a flat text sequence, we structure video reasoning chains into knowledge graphs (\textit{i.e.}, entity, relation and attribute). Within this structured framework, we inject plausible counterfactual perturbations selected based on linguistic probability distributions, creating strong misleading text that conflict with visual reality. This allows us to systematically evaluate whether mainstream LMMs can ground their reflection in visual evidence or succumb to the injected hallucinations.

Our analysis reveals that LMMs exhibit weak self-reflection capabilities. Crucially, we observe that this reflection is predominantly derived from the textual history rather than visual evidence, rendering models unable to effectively challenge more complex errors.

Intuitively, addressing textual inertia requires prompting the model to think more groundedly and removing erroneous textual history to reduce interference from textual noise.
To this end, we propose \textbf{Active Visual-Context Refinement}, a training-free inference-time strategy designed to facilitate robust self-correction. 
Emulating active visual perception, our approach actively interrupts the generation flow at key reasoning nodes to perform a look-back operation on specific video frames, thereby enforcing cross-modal interaction and ensuring the reflection is grounded in visual evidence. 
Furthermore, simply detecting an error is insufficient if the wrong tokens remain in the context window to bias future generation. 
Therefore, we introduce a folding mechanism to manage context cleanliness by compressing the trial-and-error history into a clean, factual summary once a correction is made.
This physically removes the toxic text tokens that drive text inertia and resets the attention landscape, allowing the model to perform subsequent reasoning with a clarified state.
Experimental results demonstrate that this paradigm effectively reactivates the model's self-correction capabilities, elevating explicit reflection rates from negligible levels to a substantial proportion while yielding a significant gain in overall task accuracy.
Our main contributions are summarized as follows:
\begin{itemize}[leftmargin=*,topsep=1pt,itemsep=1pt]
    \item We identify the text inertia in LMMs reasoning, revealing that models prioritize erroneous textual history over visual evidence during self-correction.
    \item We propose the LogicGraph Perturbation Protocol to systematically analyze reflection failures, uncovering that mainstream LMMs exhibit weak self-reflection capabilities, predominantly sourcing their reflection from the hallucinated textual context.
    \item We introduce Active Visual-Context Refinement, a novel inference-time strategy that integrates visual re-grounding with context denoising, significantly improving robustness and reasoning accuracy on complex video benchmarks.
\end{itemize}

\section{The LogicGraph Perturbation Protocol}
\begin{figure*}[t]
    \centering
    \includegraphics[width=1\textwidth]{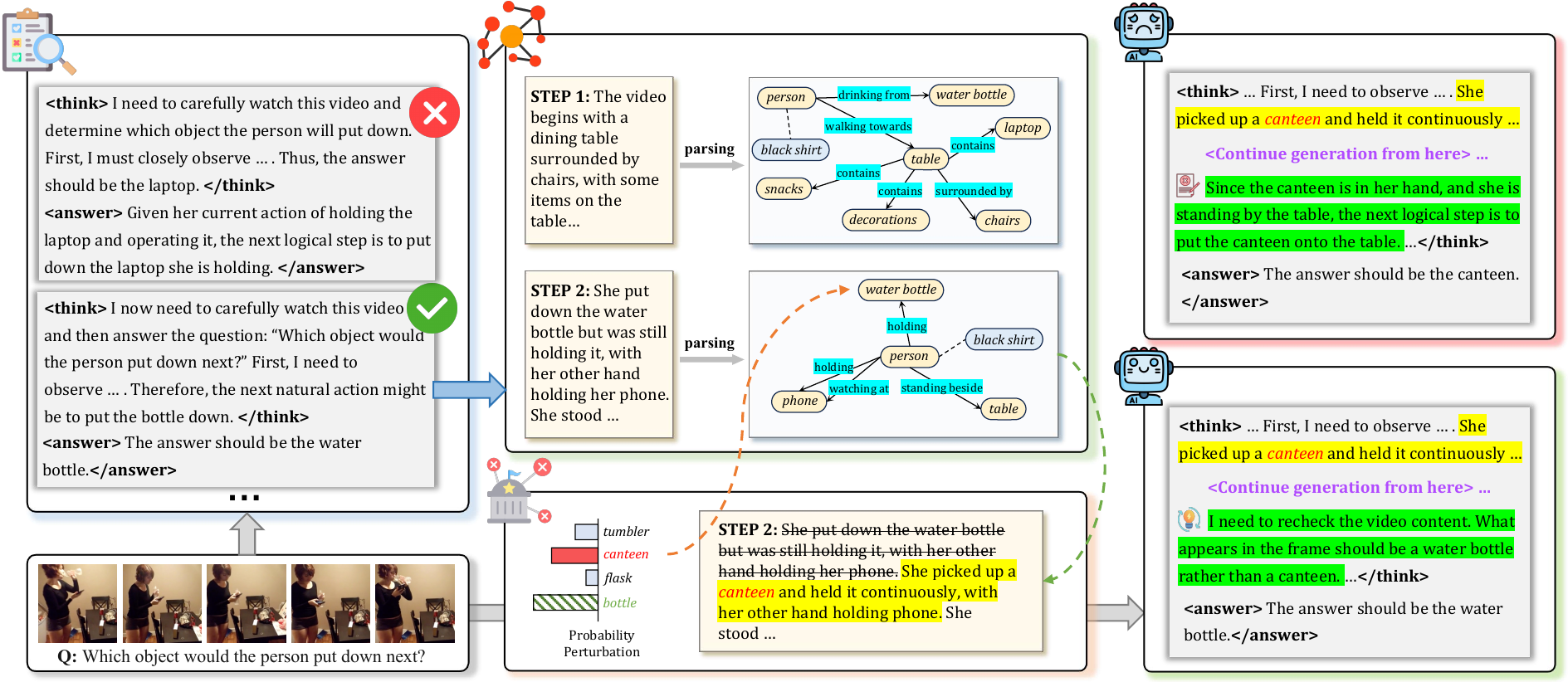}
  \caption{Overview of the \textbf{LogicGraph Perturbation Protocol}. The framework systematically evaluates text inertia by structuring reasoning chains into semantic graphs and injecting probability-weighted counterfactual perturbations. This process creates a conflict between textual priors and visual reality to determine whether the model succumbs to contextual contamination or achieves explicit reflection through visual evidence.}
    \label{figure:analysis_frame}
\end{figure*}

To transcend conventional outcome-oriented evaluations and probe the underlying cognitive dynamics of multimodal reasoning, we introduce the LogicGraph Perturbation Protocol. This framework is designed to systematically investigate the mechanism of text inertia, specifically aiming to determine whether LMMs possess the agency to rectify logic chains contaminated by textual errors through visual grounding, or if they prioritize textual consistency over visual fidelity. The overall pipeline is illustrated in Figure~\ref{figure:analysis_frame}.

\subsection{Dataset Curation}
\begin{figure}[t]
    \centering
    \begin{minipage}[c]{0.58\linewidth} 
        \small
        \setlength{\tabcolsep}{2pt} 
        \renewcommand\arraystretch{1.1} 
        
        \begin{tabular}{l r}
            \toprule
            \textbf{Category} & \textbf{Size} \\
            \midrule
            Initial Pool Size & 590 \\
            ~- Feasibility Samples & 490 \\
            ~- Prediction Samples & 100 \\
            \midrule
            Video Sources & 100 \\
            ~- Maximum Duration & 38.3s\\
            ~- Minimum Duration & 3.1s\\
            ~- Average Duration & 7.82s\\
            \bottomrule
        \end{tabular}
    \end{minipage}
    \hfill 
    \begin{minipage}[c]{0.40\linewidth}
        \centering
        
        \includegraphics[width=\linewidth]{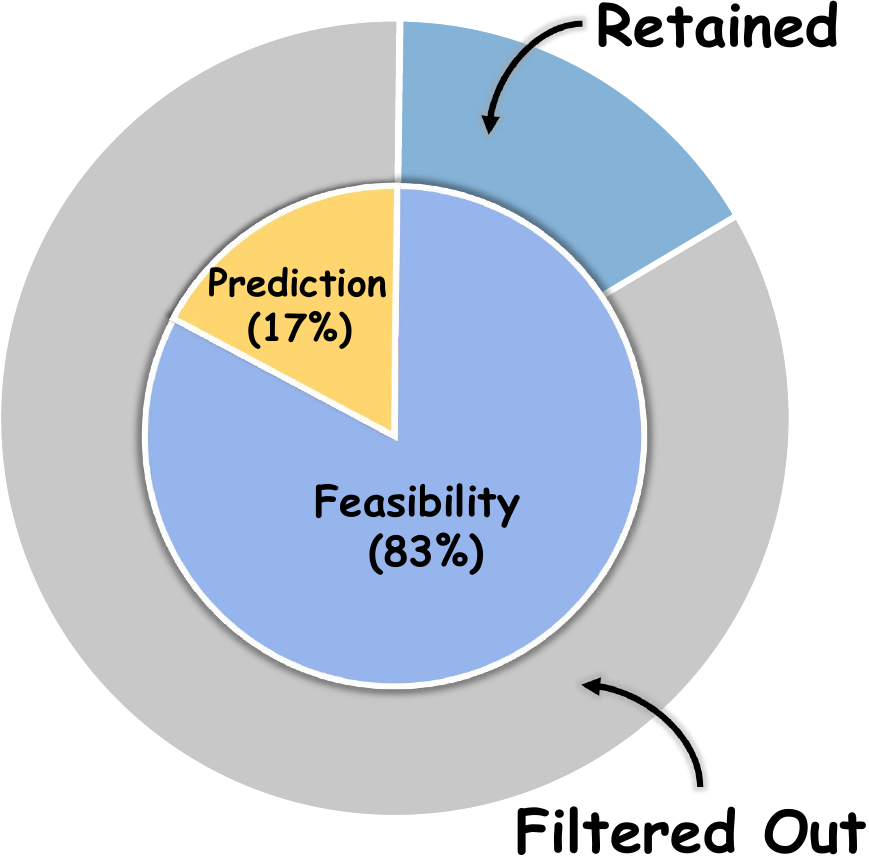}
    \end{minipage}
    
    \caption{Statistics of the curated dataset derived from STAR, showing the distribution of task types and video properties across 100 high-quality samples.}
    \label{fig:statistics}
\end{figure}

We utilize the STAR dataset~\cite{STAR}, specifically focusing on feasibility and prediction tasks. Unlike standard captioning benchmarks, this dataset requires models to perform logical deduction across temporal sequences rather than mere visual recognition. To prevent models from exploiting elimination strategies inherent in multiple-choice questions, we reformulate the dataset into an Open-ended QA format. This modification compels the model to generate explicit, self-contained reasoning trajectories, which are essential for our subsequent graph-based structural analysis. To ensure the validity of our adversarial targets, we implement a strict consistency filtering process. We retain only those samples where the reasoning trace is logically consistent with both the final answer and the ground truth. From the initial pool, we curate a high-quality subset of 100 samples, with detailed statistics and distributions presented in Figure~\ref{fig:statistics}.\looseness=-1

\subsection{Graph-based Reasoning Structuring}

To inject precise perturbations, we must first structure the unstructured text generation. We first refine the raw reasoning chains to eliminate redundancy and filler tokens, distilling the core logic. We then decompose this condensed chain into discrete reasoning steps $S = \{s_1, s_2, ..., s_n\}$. Utilizing GPT-4o as a semantic parser, we extract a semantic graph tuple $G_i = \langle E, R, A \rangle$ for each step $s_i$, where $E$ represents Entities, $R$ represents Relations (temporal or spatial), and $A$ represents Attributes. This structuring allows us to target specific logical atoms rather than arbitrarily perturbing tokens.

\subsection{Probability-Weighted Perturbation Injection}

To ensure perturbations are linguistically coherent and effectively trigger text inertia, we prioritize natural errors. For a target element $g \in \{E, R, A\}$, we generate contextually plausible but visually incorrect candidates $C$ using GPT-4o. We select the candidate $c^*$ that maximizes the joint linguistic probability of both the term and the surrounding context:
\begin{equation}
\label{equation:weighted_perturbation}
c^* = \operatorname*{argmax}_{c \in C} \frac{1}{2} \left( \mathcal{P}_{\text{token}(c)} + \mathcal{P}_{\text{sentence}(c)} \right),\nonumber
\end{equation}
where $\mathcal{P}_{\text{token}}$ and $\mathcal{P}_{\text{sentence}}$ denote the average log-probabilities of the candidate tokens and the complete sentence sequence, respectively, computed by the target LMM $P_M$ given history $H$. This selection strategy identifies the maximum likelihood hallucination, creating a plausible trap tailored to the model's distribution.

\begin{table*}[!t]
    \centering
    \small
    \begin{tabular}{l ccccc @{\hskip 0.15in} ccccc @{\hskip 0.15in} ccccc}
    \toprule
    \multirow{2}{*}{\textbf{Model}} & \multicolumn{5}{c}{\textbf{Entity}} & \multicolumn{5}{c}{\textbf{Attribute}} & \multicolumn{5}{c}{\textbf{Relation}} \\
    \cmidrule(lr){2-6} \cmidrule(lr){7-11} \cmidrule(lr){12-16}
     & Acc & $R_0$ & $R_1$ & $R_2$ & $R_3$ & Acc & $R_0$ & $R_1$ & $R_2$ & $R_3$ & Acc & $R_0$ & $R_1$ & $R_2$ & $R_3$ \\
    \midrule
    
    \multicolumn{16}{l}{\textit{\textbf{Step: First}}} \\
    \addlinespace[3pt]
    Keye-Prev-8B & 36.0 & 90.0 & 7.0 & \textbf{3.0} & 0.0 & 61.0 & 72.0 & 28.0 & 0.0 & 0.0 & 47.0 & 84.0 & 16.0 & 0.0 & 0.0 \\
    Keye-1.5-8B & \textbf{50.0} & 92.0 & 7.0 & 1.0 & 0.0 & \textbf{70.0} & 72.0 & 27.0 & \textbf{1.0} & 0.0 & \textbf{64.0} & 85.0 & 14.0 & \textbf{1.0} & 0.0 \\
    LongVILA-7B & 41.0 & 82.0 & 16.0 & 2.0 & 0.0 & 53.0 & 53.0 & 47.0 & 0.0 & 0.0 & 40.0 & 74.0 & 25.0 & \textbf{1.0} & 0.0 \\
    InternVL3-8B & 42.0 & 76.0 & 22.0 & 2.0 & 0.0 & 60.0 & 53.0 & 47.0 & 0.0 & 0.0 & 46.0 & 79.0 & 21.0 & 0.0 & 0.0 \\
    Qwen2.5-VL-7B & 28.0 & 79.0 & 21.0 & 0.0 & 0.0 & 48.0 & 50.0 & 49.0 & 0.0 & 1.0 & 42.0 & 71.0 & 28.0 & 0.0 & 1.0 \\
    \midrule
    
    \multicolumn{16}{l}{\textit{\textbf{Step: Second}}} \\
    \addlinespace[3pt]
    Keye-Prev-8B & 35.0 & 89.0 & 10.0 & \textbf{1.0} & 0.0 & 67.0 & 73.0 & 25.0 & \textbf{2.0} & 0.0 & 60.0 & 73.0 & 24.0 & \textbf{3.0} & 0.0 \\
    Keye-1.5-8B & \textbf{50.0} & 93.0 & 7.0 & 0.0 & 0.0 & \textbf{78.0} & 72.0 & 28.0 & 0.0 & 0.0 & \textbf{72.0} & 74.0 & 26.0 & 0.0 & 0.0 \\
    LongVILA-7B & 33.0 & 88.0 & 11.0 & \textbf{1.0} & 0.0 & 61.0 & 62.0 & 36.0 & 1.0 & 1.0 & 43.0 & 72.0 & 27.0 & 1.0 & 0.0 \\
    InternVL3-8B & 41.0 & 86.0 & 12.0 & \textbf{1.0} & 1.0 & 65.0 & 60.0 & 39.0 & 1.0 & 0.0 & 57.0 & 75.0 & 23.0 & 2.0 & 0.0 \\
    Qwen2.5-VL-7B & 35.0 & 86.0 & 14.0 & 0.0 & 0.0 & 55.0 & 53.0 & 47.0 & 0.0 & 0.0 & 49.0 & 65.0 & 34.0 & 1.0 & 0.0 \\
    \midrule
    
    \multicolumn{16}{l}{\textit{\textbf{Step: Third}}} \\
    \addlinespace[3pt]
    Keye-Prev-8B & 44.0 & 82.0 & 15.0 & 2.0 & 1.0 & 58.0 & 68.0 & 32.0 & 0.0 & 0.0 & 64.0 & 62.0 & 35.0 & 3.0 & 0.0 \\
    Keye-1.5-8B & 53.0 & 85.0 & 13.0 & 2.0 & 0.0 & 66.0 & 70.0 & 30.0 & 0.0 & 0.0 & 75.0 & 69.0 & 30.0 & 1.0 & 0.0 \\
    LongVILA-7B & \textbf{71.0} & 85.0 & 8.0 & \textbf{7.0} & 0.0 & \textbf{74.0} & 61.0 & 36.0 & \textbf{2.0} & 1.0 & \textbf{81.0} & 67.0 & 29.0 & \textbf{4.0} & 0.0 \\
    InternVL3-8B & 66.0 & 83.0 & 12.0 & 5.0 & 0.0 & 73.0 & 61.0 & 38.0 & 1.0 & 0.0 & \textbf{81.0} & 71.0 & 27.0 & 2.0 & 0.0 \\
    Qwen2.5-VL-7B & 48.0 & 86.0 & 12.0 & 2.0 & 0.0 & 65.0 & 64.0 & 35.0 & 1.0 & 0.0 & 56.0 & 66.0 & 31.0 & 3.0 & 0.0 \\
    
    \bottomrule
    \end{tabular}
    \caption{Quantitative evaluation of reflection capabilities across varying perturbation steps and domains. Metrics include Task Accuracy (Acc) and the distribution of reasoning behaviors ($R_0$: Contamination, $R_1$: Passive, $R_2$: Explicit, $R_3$: Collapse). \textbf{Bold} indicates the best result.}
    \label{tab:main_results}
\end{table*}

  \label{section:pipeline}
\section{Evaluations}
To systematically investigate the intrinsic reflection capabilities of LMMs and identify whether their reasoning is grounded in visual evidence or textual context, we conduct a comprehensive evaluation using the LogicGraph Perturbation Protocol proposed in \S\ref{section:pipeline}. We establish an \textit{Open-Ended Continuation} setting where the model acts as a completer: given the perturbed reasoning history, it must generate the subsequent reasoning steps to answer the question.\looseness=-1

\subsection{Models}
\begin{table*}[!t] 
    \small 

    \begin{tabular*}{\textwidth}{@{\extracolsep{\fill}}l ccccc ccccc}
    \toprule
    \multirow{2}{*}{\textbf{Model}} & \multicolumn{5}{c}{\textbf{Entity}} & \multicolumn{5}{c}{\textbf{Attribute}} \\
    \cmidrule(lr){2-6} \cmidrule(lr){7-11}
     & Acc & $R_0$ & $R_1$ & $R_2$ & $R_3$ & Acc & $R_0$ & $R_1$ & $R_2$ & $R_3$ \\
    \midrule
    
    \multicolumn{11}{@{}l}{\textit{\textbf{Step: First}}} \\
    \addlinespace[3pt] 
    
    Keye-Prev-8B & \res{52.6}{15.8} & 100.0 & 0.0 & 0.0 & 0.0 & \res{66.7}{-6.7} & \res{73.3}{-20.0} & \res{26.7}{20.0} & 0.0 & 0.0 \\
    
    Keye-1.5-8B & 47.4 & 94.7 & 0.0 & 5.3 & 0.0 & 66.7 & 80.0 & 20.0 & 0.0 & 0.0 \\
    
    LongVILA-7B & \res{47.4}{10.5} & 89.5 & \res{5.3}{-5.3} & \res{5.3}{5.3} & 0.0 & \res{80.0}{13.3} & \res{73.3}{-6.7} & 20.0 & 0.0 & \res{6.7}{6.7} \\
    
    InternVL3-8B & \res{52.6}{-5.3} & \res{84.2}{5.3} & \res{15.8}{-5.3} & 0.0 & 0.0 & 73.3 & \res{73.3}{-6.7} & \res{26.7}{6.7} & 0.0 & 0.0 \\
    
    Qwen2.5-VL-7B & 26.3 & 84.2 & 15.8 & 0.0 & 0.0 & 60.0 & \res{46.7}{-13.3} & \res{53.3}{13.3} & 0.0 & 0.0 \\
    \midrule
    
    \multicolumn{11}{@{}l}{\textit{\textbf{Step: Second}}} \\
    \addlinespace[4pt]
    
    Keye-Prev-8B & \res{48.1}{14.8} & 96.3 & 3.7 & 0.0 & 0.0 & \res{72.7}{9.1} & \res{90.9}{-9.1} & \res{9.1}{9.1} & 0.0 & 0.0 \\
    
    Keye-1.5-8B & 44.4 & 96.3 & 3.7 & 0.0 & 0.0 & 81.8 & 81.8 & 18.2 & 0.0 & 0.0 \\
    
    LongVILA-7B & \res{44.4}{3.7} & \res{85.2}{-11.1} & \res{7.4}{3.7} & \res{7.4}{7.4} & 0.0 & 54.5 & 81.8 & 18.2 & 0.0 & 0.0 \\
    
    InternVL3-8B & \res{37.0}{-3.7} & \res{81.5}{-3.7} & \res{14.8}{3.7} & \res{3.7}{3.7} & 0.0 & \res{54.5}{-9.1} & \res{63.6}{-9.1} & \res{36.4}{9.1} & 0.0 & 0.0 \\
    
    Qwen2.5-VL-7B & \res{51.9}{-3.7} & \res{74.1}{-22.2} & \res{25.9}{22.2} & 0.0 & 0.0 & 63.6 & \res{63.6}{9.1} & \res{36.4}{-9.1} & 0.0 & 0.0 \\
    
    \bottomrule
    \end{tabular*}
    \caption{Impact of decreasing perturbation strength across the first two reasoning steps. Subscripts indicate the absolute performance change compared to the high-perturbation baseline.}
    \label{tab:ablation_strength}
\end{table*}

We evaluate a diverse set of LMMs spanning both native reasoning architectures, such as Keye-preview-8B~\cite{Keye-Pre}, Keye-1.5-8B~\cite{Keye-1_5}, and LongVILA-R1-7B~\cite{Longvila}, and prompt-driven paradigms, including InternVL3-8B~\cite{InternVL3} and Qwen2.5-VL-7B~\cite{Qwen2.5-VL}. This selection allows us to comprehensively assess reflection capabilities across varying model designs. All experiments utilize a pass@3 setting. \looseness=-1

\subsection{Evaluation Metrics}

\noindent{\textbf{Task Accuracy (Acc):}}\quad
We evaluate the fundamental correctness of the final answer. Let $y$ denote the ground truth and $\hat{y}$ be the model's prediction. The accuracy is calculated as the proportion of correct matches: $\text{Acc} = \frac{1}{N} \sum_{i=1}^{N} \mathbb{I}(\hat{y}_i = y_i)$, regardless of the intermediate reasoning path.

\noindent\textbf{Reasoning Behavior Analysis.}\quad
To rigorously audit the reasoning trajectory, we classify the model's response to perturbations into four distinct categories and report their respective rates ($R_k$):

\noindent\textbf{Contextual Contamination ($R_0$):}
This signifies a visual grounding failure where reasoning is corrupted by the injected error. It manifests primarily as Direct Acceptance, where the model incorporates the perturbation $c^*$ as fact, or Rationalization, where it hallucinates visual details to logically justify the presence of the erroneous entity.

\noindent\textbf{Passive Reflection ($R_1$):}
The model derives a correct answer aligned with visual evidence but completely bypasses the textual conflict. It treats the perturbed text as absent, neither adopting nor refuting it. This reveals a critical insensitivity to contradictions, failing to explicitly resolve the cross-modal discrepancy.

\noindent\textbf{Explicit Reflection ($R_2$):}
The ideal behavior where the model actively detects the discrepancy and explicitly refutes the textual error using visual evidence. This demonstrates the capacity to override strong textual priors with veridical visual data for robust self-correction.

\noindent\textbf{Reasoning Collapse ($R_3$):} 
The injection of perturbations triggers a breakdown in the decoding process, manifesting as repetitive loops or incoherent text. This serves as a proxy for evaluating inference stability under strong cross-modal conflicts.

\noindent{\textbf{Implementation Details.}}\quad
We process video inputs by sampling frames at 5.0 fps. For each query, we sample $k=3$ reasoning chains using a temperature of 0.7. We utilize an LLM-based judge (\texttt{Qwen2.5-72B-Instruct-GPTQ-Int8}) to parse the generated outputs into the behavioral categories defined above. The final metric for each sample is aggregated based on the majority vote of the three trails to ensure robust evaluation.

\subsection{Main Results}

We present the quantitative results of reflection capabilities across all evaluated models in Table~\ref{tab:main_results}.

\textit{\textbf{Accuracy and Reflection.}}
As observed in the results, the task accuracy of all models experiences varying degrees of degradation under perturbation.
Beyond the general accuracy degradation, a more critical finding is that the \textit{Explicit Reflection ($R_2$)} remains consistently low (<10\%) across all models. Decomposing the results reveals that most correct answers stem from \textit{Passive Reflection ($R_1$)}, implying that models largely ignore conflicts rather than actively engaging in visual re-grounding to resolve discrepancies.

\textit{\textbf{Textual Inertia and Entity Vulnerability.}}
The \textit{Contextual Contamination ($R_0$)} exceeds 60\% in most scenarios, with even native reasoning models frequently rationalizing injected errors. Notably, \textit{Entity} perturbations induce the most severe degradation compared to \textit{Attribute} and \textit{Relation} types. This suggests LMMs are particularly vulnerable to entity-level hallucinations that directly conflict with object-centric visual representations.  

\textit{\textbf{The Temporal Position Effect.}}
As the perturbation position moves later in the reasoning chain (from first step to third step), both accuracy and reflection metrics improve noticeably. We attribute this to the accumulation of correct textual priors rather than improved visual grounding, as performance is poorest at the first step where models must rely solely on vision. This dependency implies that current reasoning is driven more by textual coherence than by robust visual re-examination.

\subsection{The Impact of Perturbation Strength}

To further investigate the mechanism behind reasoning failures, we hypothesize that the repetition of erroneous tokens in the context strengthens the bias towards text over vision.

We curate a subset of samples where the erroneous token appears exactly twice in the context history ($count=2$) and manually reduce it to a single occurrence ($count=1$) to lower interference strength.
Re-evaluating models on this subset (Table~\ref{tab:ablation_strength}), we observe a consistent trend across most models and phases: reducing hallucination redundancy leads to a slight decrease in the Contextual Contamination ($R_0$) and a corresponding rise in Passive Reflection ($R_1$).
However, a significant increase in Explicit Reflection ($R_2$) is rarely observed.
This indicates that while lowering textual interference reduces direct hallucination acceptance, it fails to trigger active self-correction.
Even with minimal textual cues, models remain hypersensitive to the error, struggling to override even subtle textual hallucinations with visual evidence.

\section{Active Visual-Context Refinement}
\begin{figure*}[t]
    \centering
    \includegraphics[width=1\textwidth]{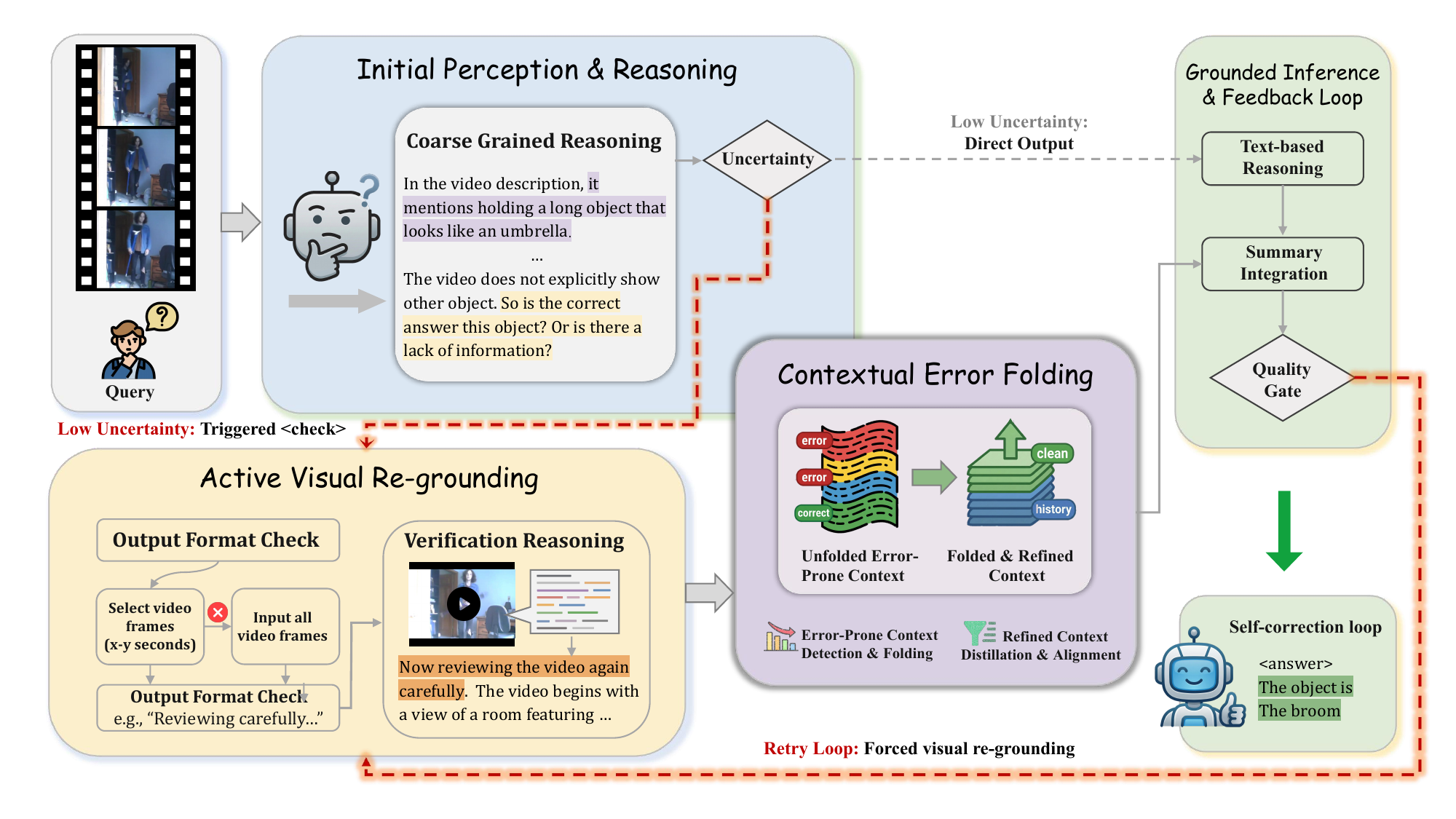}
  \caption{Overview of the Active Visual Context Refinement framework. It orchestrates an agentic loop to retrieve visual evidence upon uncertainty and folds erroneous history for robust reasoning.}
    \label{figure:method_frame}
\end{figure*}

From the perspective of multimodal cognitive alignment, current LMMs often exhibit a dependency on textual priors or generated context, overriding visual signals during complex reasoning tasks. This misalignment directs the decoder's attention mass toward the erroneous history rather than the visual tokens, leading to hallucination loops. To mitigate this issue, we propose Active Visual-Context Refinement (AVCR), a training-free framework that encourages LMMs to simultaneously enforce visual grounding and manage context cleanliness. Synthesizing insights from recent "Think-with-Image" approaches~\cite{deepeyes} and context compression strategies like ReSum~\cite{resum} and AgentFold~\cite{agentfold}, AVCR transforms the passive generation process into an agentic loop of \textit{Check}, \textit{Reason}, and \textit{Fold}.

\subsection{Problem Formulation}

In traditional Chain-of-Thought (CoT) reasoning, the generation is a static sequence where the likelihood of the next token $x_{t}$ depends primarily on the previous tokens $x_{<t}$ and the global video features $V$: $P(x_t | x_{<t}, V)$.
However, this formulation lacks the mechanism to recover from hallucinations once $x_{<t}$ contains errors.

We extend the reasoning process to an Agentic Markov Decision Process (MDP).  
Unlike standard CoT with the static visual input, our framework allows the model to dynamically alter its state.
At step $t$, the agent state $S_t$ is defined as a tuple of the current visual perception and the active textual context:
\begin{equation}
S_t = \langle \mathcal{V}t, \mathcal{C}t \rangle \nonumber
\end{equation}
where $\mathcal{V}_t$ represents the currently attended video features, and $\mathcal{C}_t$ represents the context buffer.
Adopting the ReAct paradigm~\cite{react}, the model policy $\pi_\theta(a_t | S_t)$ generates an action $a_t$, which can be a standard reasoning token or a functional token to trigger state transitions.
The AVCR framework operates through two interleaved mechanisms: Uncertainty-Driven Visual Re-grounding ($\mathcal{A}_{check}$) to update $\mathcal{V}_t$, and Context Denoising ($\mathcal{A}_{fold}$) to refine $\mathcal{C}_t$.

\subsection{Uncertainty-Driven Visual Re-grounding}

To mitigate the dominance of textual priors, the module is designed to interrupt generation when uncertainty arises. We introduce the $\mathcal{A}_{check}$ action, triggered by the \texttt{\blue{<check>}} token. Instead of attending to the entire video indiscriminately, the module predicts a specific temporal window relevant to the current reasoning node.
Upon generating \texttt{\blue{<check>}}, the model enters a decoding branch to predict a timestamp tuple $\tau = (t_{start}, t_{end})$.

\begin{equation}
V_{local} = \text{Extract}(V, \tau) \quad \text{if} \quad \text{Format}(\tau) \text{ is valid} \nonumber
\end{equation}
Emulating human visual attention by isolating critical frames, this design incorporates a feedback fallback mechanism to enhance robustness. Specifically, if the decoder fails to output a valid timestamp format, the system automatically reverts to the global video input $V_{global}$, thereby maintaining inference continuity and preventing interruptions caused by formatting anomalies.

\subsection{Context Denoising via Folding}
Even with visual re-grounding, the erroneous text tokens generated prior to correction remain in the context window. As highlighted in ReSum~\cite{resum} and analogous to memory interference mechanisms in cognitive agents~\cite{AIBrain}, these tokens act as attention sinks, biasing future generation. To address this, we introduce the Context Folding mechanism ($\mathcal{A}_{fold}$).

\noindent\textbf{The Folding Operation.}\quad
This operation is conditional: it triggers specifically when a reasoning segment concludes with a correction or a high-redundancy chain.
Instead of discarding the volatile history $H_{raw}$, which contains the error, the check action, and the correction, we synthesize a concise, factual summary $S_{fact}$ and append it to the sequence. By explicitly integrating this distilled summary into the verbose history, the mechanism effectively mitigates contextual interference. This ensures that the model prioritizes verified information over the noisy trace, preventing the accumulation of toxic reasoning paths and guiding the attention away from prior hallucinations, thereby grounding subsequent inference in corrected knowledge.\looseness=-1

We further employ a lightweight self-evaluation mechanism to audit the quality of the generated chain. The evaluator identifies signs of epistemic uncertainty or logical contradictions between the intermediate reasoning and the final answer. Upon detecting such inconsistencies, the framework triggers a global retry.
This recovery mechanism enforces mandatory visual re-grounding to ensure the final conclusion aligns with verified visual evidence.\looseness=-1

\section{Experiments}
\subsection{Experimental Setup}

\noindent{\textbf{Baselines and Models.}}\quad 
We benchmark Active Visual-Context Refinement (AVCR) against two distinct baselines using the perturbed STAR dataset constructed in \S\ref{section:pipeline}. The first baseline is Visual Focus, which explicitly directs the model via system instructions to prioritize environmental details and specific actor interactions. The second baseline is Textual Check, mimicking a look-back~\cite{Look-Back} mechanism where the model generates an initial hypothesis and performs a text-based verification of visual evidence within check tags before finalizing the answer. For these experiments, we employ KeyE-preview~\cite{Keye-Pre}, a specialist in temporal logic, and Qwen2.5-VL-7B~\cite{Qwen2.5-VL}, a general-purpose model known for robust instruction following.

\begin{table}[t]
    \centering
    \setlength{\tabcolsep}{3pt}
    
    \resizebox{\linewidth}{!}{%
    \begin{tabular}{@{}l c ccccc@{}}
    \toprule[1.5pt]
    \multirow{2}{*}{\textbf{Method}} & \multirow{2}{*}{\textbf{Step}} & \multicolumn{5}{c}{\textbf{Categories}} \\
    \cmidrule(lr){3-3} \cmidrule(l){3-7}
     & & Acc & $R_0$ & $R_1$ & $R_2$ & $R_3$ \\
    \midrule
    
    \multicolumn{7}{l}{\textit{\textbf{Model: KeyE-Preview-8B}}} \\
    \addlinespace[3pt]
    
    \multirow{2}{*}{w/ \textit{Visual Focus}} 
      & 1st & 37.0 & 89.0 & 6.0 & 5.0 & 0.0 \\
      & 2nd & 36.0 & 89.0 & 11.0 & 0.0 & 0.0 \\
    \addlinespace[2pt]
    
    \multirow{2}{*}{w/ \textit{Textual Check}} 
      & 1st & 38.0 & 82.0 & 9.0 & 9.0 & 0.0 \\
      & 2nd & 39.0 & 88.0 & 12.0 & 0.0 & 0.0 \\
    \addlinespace[2pt]
    
    \multirow{2}{*}{w/ AVCR (ours)}
      & 1st & \textbf{47.0} & 63.0 & 8.0 & \textbf{29.0} & 0.0 \\
      & 2nd & \textbf{44.0} & 70.0 & 11.0 & \textbf{19.0} & 0.0 \\
      
    \midrule
    
    \multicolumn{7}{l}{\textit{\textbf{Model: Qwen2.5-VL-7B}}} \\
    \addlinespace[3pt]
    
    \multirow{2}{*}{w/ \textit{Visual Focus}} 
      & 1st & 27.0 & 77.0 & 22.0 & 1.0 & 0.0 \\
      & 2nd & 38.0 & 81.0 & 19.0 & 0.0 & 0.0 \\
    \addlinespace[2pt]
    
    \multirow{2}{*}{w/ \textit{Textual Check}} 
      & 1st & 27.0 & 77.0 & 20.0 & 3.0 & 0.0 \\
      & 2nd & 36.0 & 78.0 & 17.0 & 5.0 & 0.0 \\
    \addlinespace[2pt]
    
    \multirow{2}{*}{w/ AVCR (ours)}
      & 1st & \textbf{36.0} & 51.0 & 18.0 & \textbf{31.0} & 0.0 \\
      & 2nd & \textbf{41.0} & 68.0 & 7.0 & \textbf{25.0} & 0.0 \\
      
    \bottomrule[1.5pt]
    \end{tabular}%
    }
    \caption{Comparison of our AVCR strategy to baseline methods on the entity perturbation domain across the first two reasoning steps.}
    \label{tab:main_exp}
\end{table}

\subsection{Main Results}

The comparative results are presented in Table~\ref{tab:main_exp}.
We find that the baseline strategies, Visual Focus and Textual Check, struggle to effectively mitigate contextual contamination.
Specifically, the \textit{Explicit Reflection ($R_2$)} remains consistently low under these settings, suggesting that mere instructional prompts are insufficient to override the strong probability mass of the hallucinated context.
While Textual Check attempts to verify the hypothesis, it often fails to ground the correction in actual visual evidence.
However, by enforcing an active perception loop and context denoising, our proposed AVCR achieves a robust improvement in both task accuracy and reflection capability, successfully overcoming the inertia that limits standard prompting approaches.

\begin{table}[t!]
    \centering
    \setlength{\tabcolsep}{3.5pt} 
    
    \resizebox{\linewidth}{!}{%
    \begin{tabular}{@{}l c ccccc@{}}
    \toprule[1.5pt]
    \multirow{2}{*}{\textbf{Method}} & \multirow{2}{*}{\textbf{Step}} & \multicolumn{5}{c}{\textbf{Categories}} \\
    \cmidrule(lr){3-3} \cmidrule(l){3-7}
     & & Acc & $R_0$ & $R_1$ & $R_2$ & $R_3$ \\
    \midrule
    
    \multicolumn{7}{l}{\textit{\textbf{Model: KeyE-Preview-8B}}} \\
    \addlinespace[3pt]
    
    \multirow{2}{*}{AVCR (ours)} 
      & 1st & \textbf{47.0} & 63.0 & 8.0 & \textbf{29.0} & 0.0 \\
      & 2nd & \textbf{44.0} & 70.0 & 11.0 & \textbf{19.0} & 0.0 \\
    \addlinespace[2pt]
    
    \multirow{2}{*}{w/o Check} 
      & 1st & 37.0 & 87.0 & 9.0 & 4.0 & 0.0 \\
      & 2nd & 36.0 & 88.0 & 10.0 & 2.0 & 0.0 \\
    \addlinespace[2pt]
    
    \multirow{2}{*}{w/o Fold} 
      & 1st & 44.0 & 67.0 & 11.0 & 22.0 & 0.0 \\
      & 2nd & 40.0 & 74.0 & 11.0 & 15.0 & 0.0 \\
      
    \midrule
    
    \multicolumn{7}{l}{\textit{\textbf{Model: Qwen2.5-VL-7B}}} \\
    \addlinespace[3pt]
    
    \multirow{2}{*}{AVCR (ours)} 
      & 1st & \textbf{36.0} & 51.0 & 18.0 & \textbf{31.0} & 0.0 \\
      & 2nd & \textbf{41.0} & 68.0 & 7.0 & \textbf{25.0} & 0.0 \\
    \addlinespace[2pt]
    
    \multirow{2}{*}{w/o Check} 
      & 1st & 28.0 & 77.0 & 23.0 & 0.0 & 0.0 \\
      & 2nd & 34.0 & 83.0 & 17.0 & 0.0 & 0.0 \\
    \addlinespace[2pt]
    
    \multirow{2}{*}{w/o Fold} 
      & 1st & 32.0 & 57.0 & 19.0 & 24.0 & 0.0 \\
      & 2nd & \textbf{41.0} & 70.0 & 9.0 & 21.0 & 0.0 \\
      
    \bottomrule[1.5pt]
    \end{tabular}%
    }
    \caption{Ablation study on different functional component regarding the entity metric across the first two reasoning steps.}
    \label{tab:ablation_exp}
\end{table}

\subsection{Impact of Key Components}

To investigate the underlying mechanisms of the improvement, we analyze the specific contribution of each component in Table~\ref{tab:ablation_exp}.
We first examine the necessity of active visual perception.
When restricted to internal textual reflection without retrieving specific video frames, the model fails to demonstrate significant improvement.
This confirms that textual reasoning alone is inadequate for resolving cross-modal discrepancies.
Furthermore, we find that accessing visual evidence is insufficient if the erroneous history persists.
When the history is retained, the model still faces interference from textual priors, which limits the efficacy of the visual correction. This motivates the context folding mechanism, which compresses the misleading history to further mitigate interference. Therefore, integrating visual re-grounding with context denoising yields the most robust self-correction.\looseness=-1

\section{Related Work}
\noindent{\textbf{Large Multimodal Models.}}\quad 
Large Multimodal Models (LMMs)~\cite{InternVL3,InternVideo2.5,VideoChat-R1} have demonstrated remarkable capabilities in long-context video understanding and temporal logic reasoning. Represented by mainstream architectures such as the InternVL series~\cite{InternVL3}, and Qwen2.5-VL~\cite{Qwen2.5-VL}, these models have evolved from handling static images to maintaining logical consistency across extensive visual streams. However, their reasoning backbone relies heavily on the decoding mechanisms of Large Language Models (LLMs), which inevitably introduces strong textual priors~\cite{Winoground,PVLP}. Consequently, preventing these models from prioritizing textual probabilities over visual reality remains a critical challenge, as this tendency frequently results in hallucination.

\noindent{\textbf{Multimodal Reflection and Hallucination.}}\quad 
Multimodal hallucination, where generated responses contradict visual content, poses a significant threat to LMM reliability~\cite{Survey_on_Hallucination,woodpecker,Prismatic_VLMs,VidHalluc,D-LEAF}. Recent advancements have leveraged reinforcement learning and long-context training to enhance the reasoning capabilities of video LMMs~\cite{videor1, Keye-Pre, Longvila}, achieving impressive performance. Despite these gains, we observe that models remain vulnerable to textual inertia~\cite{MoreAttention,ImagesSpeak,ReCoT}, where early errors in a reasoning chain bias subsequent outputs, overriding visual evidence. To address this, research has explored self-reflection mechanisms to rectify reasoning chains~\cite{Reflexion,SCoRE, ReCoT}. Distinct from previous approaches, our proposed AVCR framework addresses the root cause by simultaneously enforcing active visual re-grounding and adaptive context folding, effectively breaking the cycle of hallucination propagation.\looseness=-1

\section{Conclusion}
In this paper, we identify the critical failure mode of textual inertia in LMMs where reasoning chains are dominated by erroneous textual priors rather than visual evidence. To systematically investigate this cognitive misalignment, we introduce the LogicGraph Perturbation Protocol which structurally injects counterfactual noise into distinct reasoning stages to probe the intrinsic reflection capabilities of diverse models. Our extensive evaluations reveal that current models exhibit fragile self-correction abilities and predominantly succumb to blind error propagation. To mitigate this, we propose Active Visual-Context Refinement. This training free paradigm orchestrates active visual re-grounding and context denoising to effectively stifle hallucination propagation and enhance reasoning robustness.\looseness=-1

\section*{Limitations}
Although we construct a rigorous evaluation protocol and a novel inference strategy, our work has limitations. First, the perturbation scenarios in our protocol are currently focused on specific logical atoms including entity and attribute errors. Expanding to more complex causal or counterfactual reasoning scenarios remains a challenge for future research. Second, our proposed AVCR is an inference time strategy. While efficient, it does not fundamentally alter the internal parameters of the model to permanently fix the attention misalignment. Third, our experiments are primarily conducted on open source models due to computational constraints. Validating the scalability of our approach on larger proprietary models requires further exploration.

\bibliography{custom}

\appendix
\section{Detailed descriptions of LMMs} \label{appendix:model_details}

In this section, we provide detailed specifications of the Multimodal Large Language Models employed in our experiments.

\noindent{\textbf{Keye-preview}}~\cite{Keye-Pre} is a specialized video reasoning model optimized for temporal logic inference. It is trained on large-scale video chain-of-thought data to enhance its native capability in deducing causal relationships and temporal sequences within dynamic visual contexts.

\noindent{\textbf{Keye-1.5}}~\cite{Keye-1_5} builds upon the foundation of Keye-preview with an expanded training corpus and refined architecture. It incorporates advanced alignment strategies to better synchronize visual perception with textual reasoning, demonstrating superior performance in complex query response tasks.

\noindent{\textbf{LongVILA-R1}}~\cite{Longvila} focuses on long-context video understanding and reasoning. By utilizing efficient token compression and training on extended video sequences, it effectively manages long-term temporal dependencies and maintains consistency across lengthy reasoning chains.

\noindent{\textbf{InternVL3}}~\cite{InternVL3} integrates a powerful InternViT visual encoder with a large language model. It employs a progressive alignment strategy to achieve robust performance across diverse multimodal tasks including image captioning and video question answering.

\noindent{\textbf{Qwen2.5-VL}}~\cite{Qwen2.5-VL} is a mainstream general-purpose model constructed upon the Qwen2.5 language model and a dynamic resolution vision transformer. It utilizes the Naive Dynamic Resolution mechanism and Multimodal Rotary Positional Embedding to effectively process visual information at varying scales and durations.

\section{Details of LogicGraph Perturbation Protocol} \label{appendix:protocol_details}

We utilize GPT-4o as the semantic parsing and perturbation engine to construct the LogicGraph Perturbation dataset. The construction process involves three distinct stages utilizing specific prompts to ensure structural integrity and effectiveness.

\subsection{Graph Structuring}

To transform unstructured reasoning chains into structured representations, we instruct GPT-4o to parse the text into semantic tuples comprising entities, relations, and attributes. The specific instruction is presented in Figure~\ref{template:struct}.

\subsection{Perturbation Generation}

Based on the extracted graph structures, we generate counterfactual perturbations that maximize linguistic probability while contradicting visual facts. The prompt ensures that the generated errors are contextually plausible to effectively trigger textual inertia. This is detailed in Figure~\ref{template:perturbation}.

\subsection{Reasoning Evaluation}

To automate the behavioral analysis of LMMs under perturbation, we employ an LLM-based judge to classify the generated reasoning chains into four categories: Contextual Contamination, Passive Reflection, Explicit Reflection, and Reasoning Collapse. The classification criteria are rigorously defined in Figure~\ref{template:evaluation}.

\section{Introduction to the STAR Dataset} 
STAR dataset~\cite{STAR} comprises approximately 60K situated reasoning questions and 22K real-world video clips. The Feasibility task probes the ability to infer viable actions under specific constraints, requiring models to deduce possibilities rather than observed facts. The Prediction task evaluates forecasting plausible future actions, where models must anticipate outcomes based on masked initial video segments.

\begin{figure*}
    \centering
    
    \begin{tcolorbox}[promptstyle={Prompt Template for Semantic Graph Structuring}]
    \small 
    
    \textbf{System Prompt:} You are an expert AI assistant specialized in analyzing and restructuring reasoning processes. Your task is to convert unstructured reasoning text into well-organized steps with knowledge graphs.

    \textbf{User Prompt:} 
    \textbf{Task Overview}
    Given a solution with multiple reasoning steps, reformat it into structured steps and knowledge graphs.

    \textbf{1. Solution Filtration}
    Filter the solution to contain only valid reasoning statements based on these rules:
    - Remove reasoning statements not supporting the final conclusion.
    - Remove statements not logically related to the core conclusion.
    - Remove repetitive statements.
    - Only perform Remove operations without making additions or modifications.

    \textbf{2. Solution Parsing}
    Convert the filtered solution into a sequence of distinct reasoning steps by segmenting the original text.
    
    Segmentation and Granularity:
    - Divide the filtered solution into coherent logical units where each unit becomes a single reasoning step.
    - A reasoning step should represent a complete thought or a set of closely related observations.
    - Group consecutive sentences that describe one clear point into a single step.
    
    For each identified reasoning step:
    - The step field content: Must be the exact verbatim segment of text from the filtered solution. Preserve original content and order. Do not add interpretations or omit parts.
    - Create a Concise Step Caption: Generate a separate brief caption summarizing the core idea of the verbatim segment.
    
    Important: The number of Concise Step Captions must exactly match the number of step field contents. These captions will form the step\_overall field.

    \textbf{3. Reasoning Step Graphing}
    Translate each reasoning step into a knowledge graph including entities, relations, and attributes.
    
    - Entities: Identify main subjects or objects as definite nouns or phrases.
    - relations: Identify logical connections, interactions, and properties expressed as triplets entity1, relation, entity2. Capture detailed information including actions, spatial, temporal, causal, comparative, or conditional connections.
    - Attributes: Store specific characteristics of an entity.

    \textbf{4. Output Format}
    Present your output as a single JSON object.
    
    [
      \{
        "raw\_solution": "The initial input solution content that was processed.",
        "filtered\_solution": "The solution content after applying filtration rules from Section 1.",
        "step\_overall": "Concise Step1 Caption -> Concise Step2 Caption -> ...",
        "Parsing": [
          \{
            "step": "Exact verbatim segment for Step 1 from filtered\_solution.",
            "graph": \{
              "entities": ["entity1\_str", "entity2\_str", ...],
              "relations": [
                ["entity1\_str", "relation1\_type\_str", "entity2\_str"], ...
              ],
              "attributes": [
                \{"entity1\_str": \{"attribute\_key": "attribute\_value"\}\}, ...
              ]
            \}
          \},
          ...
        ]
      \}
    ]
    
    Here is the problem, and the solution that needs to be reformatted to steps:
    
    [Problem]
    \textcolor{myblue}{\{question\}}
    
    [Solution]
    \textcolor{myblue}{\{think\}}
    
    [Correct Answer]
    \textcolor{myblue}{\{answer\}}
    
    \end{tcolorbox}
    
    \caption{The prompt template used for Semantic Graph Structuring.}
    \label{template:struct} 
\end{figure*}

\begin{figure*}
    \centering
    
    \begin{tcolorbox}[promptstyle={Prompt Template for Perturbation Generation}]
    \small 
    
    \textbf{System Prompt:} You are an expert AI assistant specialized in introducing targeted, plausible modifications to reasoning processes. Your task is to create strategic hallucinations that can test model robustness while maintaining realistic plausibility.

    \textbf{User Prompt:} 
    Your task: Generate FIVE different targeted hallucinations by modifying the SAME entity in the FIRST Parsing step of the input JSON. Each modification should create a different incorrect conclusion.

    Input JSON fields:
    - raw\_solution: Full reasoning with final conclusion.
    - filtered\_solution: Concise reasoning steps.
    - step\_overall: Summary string (e.g., "A -> B -> C").
    - Parsing: Array of \{step: reasoning sentence(s), graph: knowledge graph\}.

    Modification Process:
    1. \textbf{Analyze}: Review the second step in Parsing and identify a key entity that significantly impacts the reasoning. You should change one entity to another, instead of modifying it into an entity with added attributes.
    2. \textbf{Select Entity}: Choose ONE entity from the first step's graph that will be modified in five different ways.
    3. \textbf{Generate Five Variations}: Create five different modifications to this same entity:
    * Each modification should change the entity to something plausible but incorrect
    * Each should lead to a different misleading conclusion
    * Maintain original reasoning style and context
    4. \textbf{Update Components}: For each variation, update the step text, graph, step\_overall, and disturbed\_raw\_solution\_prefix accordingly.

    OUTPUT JSON SPECIFICATIONS:

    Your entire response MUST be a singe, valid json object. No text/markdown outside the main \{\}.

    The JSON object MUST contain these fields:

    1. generation\_explanation: (String) Explanation of which entity was selected and how five different modifications were applied.
    2. selected\_entity: (String) The original entity name that was modified across all five variations.
    3. variations: (Array of 5 objects) Each object contains:
    \{
        "variation\_id": 1-5,
        "modified\_entity": "new entity name for this variation",
        "step\_overall": "Updated step\_overall with first step modified for this variation",
        "Parsing": [Updated Parsing array ,only need to generate first modified step and graph for this variation],
        "step\_prefix": "place the complete modified first step here"
        "disturbed\_raw\_solution\_prefix": "this string should contain the raw\_solution first text up to end of modified first complete step for this variation, for example, in this raw\_solution: xxx(there are unfiltered text) , step1, yyy(there are unfiltered text), step2 ..., you should include xxx and modified step1(not include yyy), noted that you can not modify the xxx(text before step1) part"
    \}

    \textbf{Important Notes:}
    - All five variations modify the SAME original entity but change it to five DIFFERENT new entities
    - Each variation should have its own step\_overall, Parsing, and disturbed\_raw\_solution\_prefix
    - Only the first step in Parsing should be modified; all other steps remain unchanged across all variations
    - Each variation should be plausible but lead to different incorrect reasoning paths

    Here is the input JSON:
    ```json
    \textcolor{myblue}{\{structured\_content\}}
    ```
    \end{tcolorbox}
    
    \caption{The prompt template used for Perturbation Generation (Entity Modification).}
    \label{template:perturbation} 
\end{figure*}

\begin{figure*}
    \centering
    
    \begin{tcolorbox}[promptstyle={Prompt Template for Reasoning Behavior Classification}]
    \small 
    
    \textbf{System Prompt:} You are a Logic Auditor evaluating the reasoning chain of a Video-LLM. The VLM has been fed a prompt containing a deliberate LIE (Modified Entity/Perturbation). Your objective is to detect if the VLM's reasoning has been CONTAMINATED by this Lie.

    THE ENTITIES:
    1. TRUTH (Original Entity)\textbf{:} The object/location actually in the video.
    2. LIE (Modified Entity)\textbf{:} The fake object/location injected into the text.

    Evaluation Categories (Mutually Exclusive):

    \textbf{Category 0: Contextual Contamination }
    This includes ANY level of acceptance or confusion caused by the Lie.
    - Direct Acceptance: The model accepts the Lie as truth.
    - Rationalization: The model invents a reason to justify the Lie (e.g., "It is near the [Lie], so...").
    - Epistemic Confusion: The model wavers or doubts the visual evidence because of the text (e.g., "Is it a [Truth] or a [Lie]? I'm not sure.").
    - Trace Contamination: The model generates the correct answer, BUT mentions the Lie as part of the final scenario description.

    \textbf{Category 1: Passive Reflection }
    - The model completely IGNORES the Lie. It describes the Truth as if the Lie was never mentioned.
    - It does NOT correct the user. It effectively treats the injected text as invisible noise.
    - Crucially: It shows NO sign of confusion and NO trace of the Lie in its final answer.

    \textbf{Category 2: Explicit Reflection }
    - The model ACTIVELY REJECTS the Lie based on visual evidence.
    - It contains phrases like: "The prompt mentions [Lie], but the video clearly shows [Truth]" or "Unlike the text suggests..."
    - It corrects the reasoning path explicitly.

    \textbf{Category 3: Reasoning Collapse }
    - Repetitive loops, incoherent text, or empty responses.

    \textbf{Response Format (JSON ONLY):}
    \{
      "contamination\_check": \{
        "direct\_mention": "Does the model explicitly mention seeing the [Lie]? (Yes/No)",
        "justification\_trace": "Does the model use the [Lie] to explain a location or action? (Yes/No)",
        "confusion\_indicator": "Does the model express doubt/hesitation between Truth and Lie? (Yes/No)"
      \},
      "reflection\_check": "Does the model explicitly state the text/prompt is wrong? (Yes/No)",
      "final\_verdict": "Category 0, 1, 2, or 3",
      "category": 0,
      "reasoning": "Concise explanation of why. If Category 0, explain if it was Direct Acceptance, Justification, or Confusion."
    \}

    Only output valid JSON, do not include any other text.

    \textbf{User Prompt:} 
    Task Data:
    - TRUTH (Video shows)\textbf{:} \textcolor{myblue}{\{selected\_entity\}}
    - LIE (Text Injected)\textbf{:} \textcolor{myblue}{\{modified\_entity\}}

    Corrupted Context given to Model:
    \textcolor{myblue}{\{partial\_input\}}

    Model's Output to Evaluate:
    \textcolor{myblue}{\{continued\_output\}}

    Audit Instructions:
    1. Read the <answer> block first\textbf{.} Does it mention the LIE (\{modified\_entity\})? If yes, even as background context, this is Category 0.
    2. Read the <think> block.
    - If the model asks "Is it \textcolor{myblue}{\{selected\_entity\}} or \textcolor{myblue}{\{modified\_entity\}}?", this is Category 0 (Confusion).
    - If the model assumes the Lie is true to make sense of the scene, this is Category 0.
    3. Only assign Category 1 if the model talks about \textcolor{myblue}{\{selected\_entity\}} 100\% confidently and never acknowledges the \textcolor{myblue}{\{modified\_entity\}} exists in the text.
    4. Only assign Category 2 if there is an explicit "No" or "Correction" regarding the text.

    \textbf{Provide the JSON audit:}
    
    \end{tcolorbox}
    
    \caption{The prompt template used for Reasoning Behavior Classification. (Entity Modification).}
    \label{template:evaluation} 
\end{figure*}

\end{document}